\documentclass{article}

\usepackage{microtype}
\usepackage{graphicx}
\usepackage{subfigure}
\usepackage{booktabs} 

\usepackage{hyperref}

\usepackage{fancyhdr}
\usepackage[accepted]{icml2023}

\usepackage{amsfonts} 
\usepackage{mathtools} 
\usepackage{amsmath}
\usepackage{amssymb}
\usepackage{mathtools}
\usepackage{amsthm}

\usepackage[capitalize,noabbrev]{cleveref}

\theoremstyle{plain}

\theoremstyle{definition}

\theoremstyle{remark}

\usepackage[textsize=tiny]{todonotes}

\icmltitlerunning{RL-Control Policy for Molecular Dynamics Simulators}

\begin{document}

\twocolumn[
\icmltitle{Augmenting Control over Exploration Space in Molecular Dynamics Simulators to Streamline De Novo Analysis through Generative Control Policies}



\icmlsetsymbol{equal}{*}

\begin{icmlauthorlist}
\icmlauthor{Paloma Gonzalez-Rojas}{mcsc}
\icmlauthor{Andrew Emmel}{bio}
\icmlauthor{Louis Martinez}{eecs}
\icmlauthor{Neil Malur}{eecs}
\icmlauthor{Gregory Rutledge}{cheme}
\end{icmlauthorlist}

\icmlaffiliation{mcsc}{Department of Chemical Engineering, MIT Climate and Sustainability Consortium, Massachusetts Institute of Technology, Cambridge, United States}
\icmlaffiliation{cheme}{Department of Chemical Engineering, Massachusetts Institute of Technology}
\icmlaffiliation{eecs}{Department of Electrical Engineering and Computer Science, Massachusetts Institute of Technology}
\icmlaffiliation{bio}{Department of Biological Engineering, Massachusetts Institute of Technology}

\icmlcorrespondingauthor{Paloma Gonzalez-Rojas}{palomagr@mit.edu}

\icmlkeywords{Machine Learning, ICML}

\vskip 0.3in
]



\printAffiliationsAndNotice

\begin{abstract}
This study introduces the P5 model - a foundational method that utilizes reinforcement learning (RL) to augment control, effectiveness, and scalability in molecular dynamics simulations (MD). Our innovative strategy optimizes the sampling of target polymer chain conformations, marking an efficiency improvement of over $37.1\%$. The RL-induced control policies function as an inductive bias, modulating Brownian forces to steer the system towards the preferred state, thereby expanding the exploration of the configuration space beyond what traditional MD allows. This broadened exploration generates a more varied set of conformations and targets specific properties, a feature pivotal for progress in polymer development, drug discovery, and material design. Our technique offers significant advantages when investigating new systems with limited prior knowledge, opening up new methodologies for tackling complex simulation problems with generative techniques.

\begin{figure}
    \centering
    \includegraphics[width=0.45\textwidth]{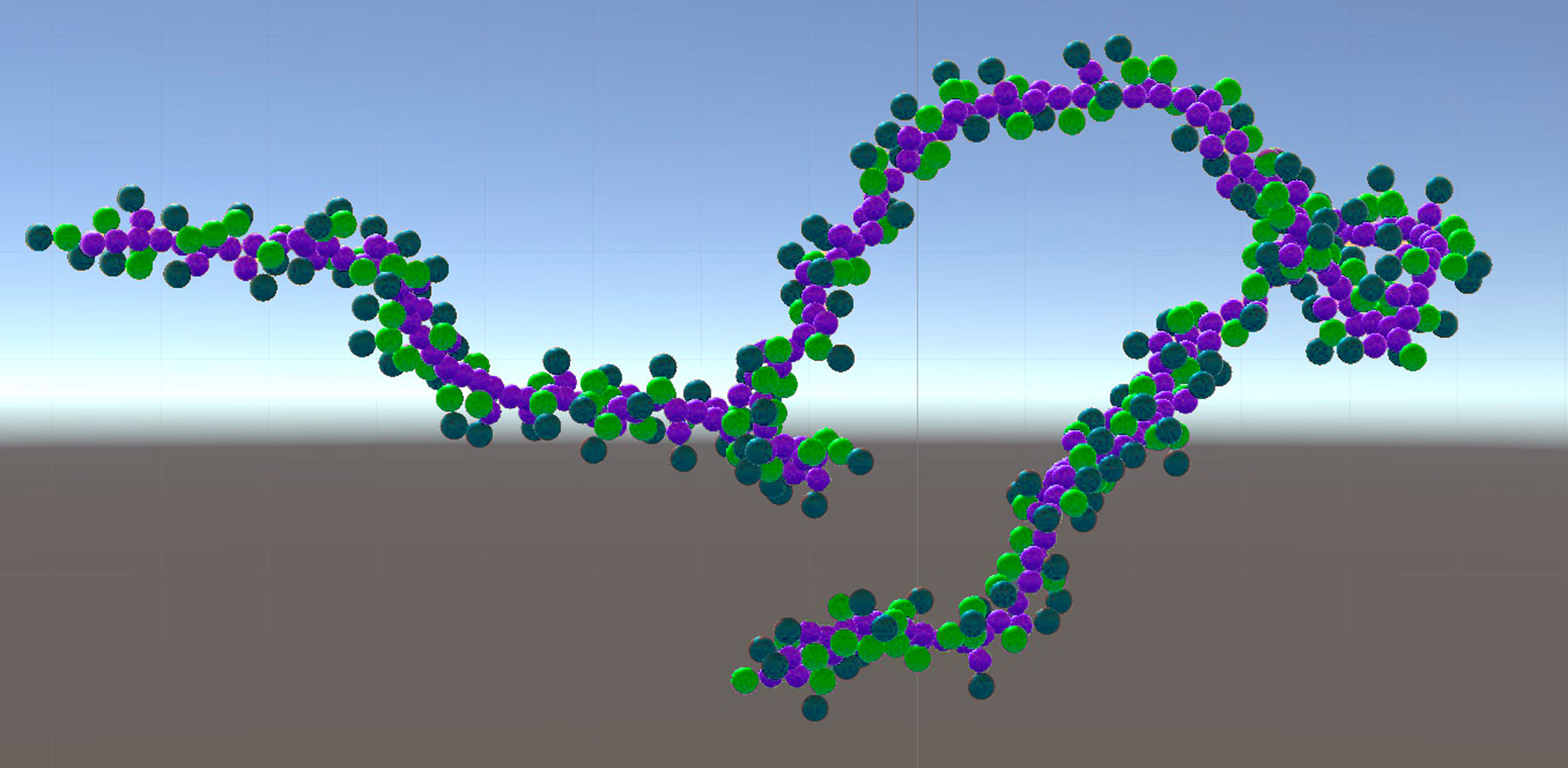}
    \caption{Simulated polymer chain of Cellulose Acetate with P5.  }
\end{figure}

\end{abstract}

\section{Introduction}
\label{submission}

Molecular dynamics has emerged as an invaluable tool for examining molecular systems, albeit with certain limitations. These methods often prove computationally intensive, particularly when handling complex systems such as macromolecules, for instance, polymer chains with more than $10^4$ atoms. In addition, uncertainties may arise due to this increasing system complexity, inaccuracies in complex definitions such as force fields, or limited knowledge about de Novo molecular systems. Machine learning has shown promise in addressing these challenges with data-driven methods like property predictions using discriminative techniques \cite{karuth_predicting_2021, tao_benchmarking_2021}, and machine learning force fields \cite{doerr_torchmd_2021, kleinschmidt_computational_2022, unke_machine_2021}.  Computational chemistry has seen machine learning techniques gradually incorporate 3D \cite{ganea_geomol_2021} and graph representations \cite{antoniuk_representing_2022, park_prediction_2022, you_graph_2018, zuo_accelerating_2021}. Recent work has included physics analysis to inform generative molecular machine learning \cite{liu_genphys_2023}. Despite these advances, deriving molecular properties that demand expectations over an ensemble of states or time-dependent behavior of atoms and molecules remains relatively unexplored.

This paper presents P5, an RL-based control policy system in a three-dimensional physics engine simulator to enhance molecular simulation processes. P5 stands for Predicting Polymer Properties and Processability with Physics-Informed Reinforcement Learning. By utilizing reinforcement learning with a molecular dynamics module, including Brownian Dynamics, we extend the scope of application for control policies in molecular dynamics simulations. Our approach demonstrated increased control over uncertainty and addressed previous limitations in modeling and manipulating the polymer chain's structural dynamics. 

Our work builds upon existing molecular dynamics sampling machine learning algorithms. While \citet{holdijk_path_2022} demonstrate how a physics-inspired approach for transition sampling may provide direct control over molecular systems, P5's RL approach offers flexibility, adaptability, and environmental feedback. \citet{klein_timewarp_2023} introduce 'TimeWarp,' which utilizes the Metropolis-Hastings algorithm for probabilistic sampling, while P5 actively targets specific states. Rose et al. \cite{rose_reinforcement_2021} deploy RL to sample rare events in their molecular dynamics research and P5 seeks an array of targets. Our approach focuses on increasing control and efficiency over macromolecule dynamics. Furthermore, it builds upon RL extensive applications in robotics through training on 3D physics engines \cite{muratore_robot_2022, brockman_openai_2016, todorov_mujoco_2012, juliani_unity_2020}, which are widely used for model-based control learning. Reinforcement learning algorithms, notably A3C \cite{mnih_asynchronous_2016} and PPO \cite{schulman_proximal_2017}, form the foundation of these platforms, driving control systems for kinetic simulators.

Through simulating a molecule of Cellulose Acetate (renewable biopolymer chain) with P5, we demonstrate how RL-based control policies offer improved exploration of the configuration space, adaptation to specific objectives, and optimization of computational resources in over a $37.1\%$  improvement of polymer chain conformation sampling targeting a known radius of gyration range. This approach boosts the efficiency of the simulation process and fosters the development of RL-based control policies, which enhance the precision and efficiency of the simulations. Furthermore, P5 can be extended to various molecular systems, underscoring its transferability to de Novo molecules, applicability to cross-scale problems, and generating new conformations. The P5 model development with inexpensive computation and data makes P5 a promising tool for accelerating polymer research and development. Furthermore, this simulation technology advancement can accelerate discoveries in various scientific and engineering domains. 

\section{Methods}

\subsection{Simulation Architecture}
We introduced P5, a generative simulator developed to predict polymer properties and processing using physics-informed reinforcement learning to advance molecular dynamics simulations. This system was built on Unity 3D, a physics game engine that allows for the simulation of Newtonian physics of motion and 3D geometric representation of chemical structures. We incorporated the ML-agents module within the Unity environment, including standard reinforcement learning algorithms, Proximal Policy Optimization (PPO) \cite{schulman_proximal_2017}, and other similar techniques. We selected PPO as our reinforcement learning algorithm due to its flexibility and balanced exploration-exploitation trade-off. We 'froze' the components of the MD, 3D-physics, RL, and the molecular system, thus concentrating our efforts on designing control policies for molecular dynamics\footnote{See Appendix 1, including error detection.}.

Our simulation framework incorporated the extensively-validated Martini Force Field \cite{bereau_automated_2015} for representing organic molecules explained in Appendix 3. The geometric modeling of P5 is performed in Unity, with hinge joints employed to emulate bonds between polymer chain beads, facilitating precise rotational dynamics and an accurate depiction of the structural intricacies of the polymer chain. We utilized SMILE and Gromacs readers for monomer topology, enabling precise bead position retrieval in Cartesian space using RDKit \cite{landrum_rdkit_2010}. The nonbonded interactions were modeled through Lennard-Jones and Brownian forces as an implicit solvent. Collectively, these methodologies contribute to a high-fidelity representation of polymeric molecular systems.

In the P5 model, the operational timescale is denoted as a dimensionless timestep, attributable to its implementation in the physics engine. This dimensionless timestep can be converted to physical time units such as femtoseconds using characteristic values of the relevant properties (e.g., bond lengths, bond angles, atomic masses) to derive the appropriate scaling factor. The selection of the exact scaling factor is contingent on the specific system under study and forms a crucial part of the model calibration process \footnote{See Appendix 2.}. 

The development and training phase of the P5 model was conducted on a standard laptop, demonstrating its compatibility with commonly accessible computing resources. However, the model can be executed on cloud computing platforms for increased performance by 10x and scalability.

\subsection{Control Policy Development}

The control policy is optimized by a neural network (NN) agent that manipulates Brownian forces acting on the backbone beads of the polymer chain (PC). This process propels the beads to swiftly alter their positions, enabling a transition from an initial conformation to a target ensemble defined by a pre-specified radius of gyration (RG) range.

\textbf{States $S_t$: }The state space, denoted by $S$, combines several elements: the Geometry component, $S_{\text{geometry}}$, dihedral angles $dihedral_s$, and bond angles $bond_s$ are calculated based on the position vectors. The Dynamics component, $S_{\text{dynamics}}$, captures the position $r_s$, velocity $v_s$, and angular velocity $w_s$ of the rigid body of the beads. The Environmental component, $S_{\text{environment}}$, reflects the positions of other neighboring beads' rigid bodies, encapsulated in the set $others_s$, within a specific observation radius around the current rigid body, to provide environmental feedback. The totality of the state space $S$ is thus a combination of these three components: $S = S_{\text{geometry}} \cup 
S_{\text{dynamics}} \cup S_{\text{environment}}$.

In the state space $S_t$, there will be a series of states corresponding to target conformations with the desired radius of gyration. These states capture the specific geometric and dynamic properties of the polymer chain. The radius of gyration (RG) is calculated in equation (1):
\[
\text{RG} = \sqrt{\frac{1}{N}\sum_{i=1}^{N}(r_i - r_{\text{cm}})^2} \tag{1}
\]
Where $N$ is the number of beads in the chain, $r_i$ represents the position of each bead, and $r_{\text{cm}}$ is the position of the center of mass of the chain, quantifying the spatial extent of the polymer chain. The RG is an intrinsic characteristic of a polymer, a known property for widely studied polymers, and obtained from the literature for P5. Its implementation established a clear target state for developing a proof of concept. In further P5 development, the RG can be estimated from the simulation data to increase generalization. In addition, the objective function is flexible enough to accommodate a broad array of target objectives.

\textbf{Actions $A_t$: }Our system's neural network (NN) is the agent that controls the Brownian dynamics by   adding a random "kick" or disturbance to the particles at each timestep, simulating the effect of random collisions with a fluid medium, and rotations to the backbone beads of each monomer in the chain, influencing the system's dynamics. The force applied to the bead is denoted as $\mathbf{F}_ { learned}$ and is calculated as follows:
$\mathbf{F}_{ learned} = \begin{bmatrix} \text{x}_a \ \text{y}_a \ \text{z}_a \end{bmatrix} \cdot (\text{coef} \cdot \alpha_a)$. Here, $\text{x}_a$, $\text{y}_a$, and $\text{z}_a$ represent the directional components of the force, while $\text{coef}$ and $\text{$\alpha$}_a$ are factors to control force magnitude. The rotation angle action is denoted as $\theta_{\text{rotation}}$ and is calculated as $\theta_{\text{rotation}} = \text{angle}_a \cdot \theta{\text{max}}$. Here, $\text{angle}_a$ represents the rotation angle action provided by the NN agent, and $\theta_{\text{max}}$ denotes the maximum rotation angle defined as a parameter of the system. The rotation angle $\theta_{\text{rotation}}$ influences the dihedral angles of the beads and ultimately affects the overall conformation of the polymer chain.

\textbf{Rewards $R_t$: }The reward function consists of three components that are functions of the radius of gyration (RG) and its deviation from the target range $targetRG_{min}$ and $targetRG_{max}$. The distance reward $R_{\text{dist reward}}$ penalizes the agent in proportion to the squared distance between the current radius of gyration and the target range. The radius of gyration reward $R_{\text{RG reward}}$, this component incentivizes the agent to maintain the radius of gyration within the target range. The third reward is the shaping reward $R_{\text{RG shaping reward}}$, which guides the agent's exploration and encourages stability. It rewards the agent for maintaining a radius of gyration near the middle of the target range and penalizes deviations.

With this reward structure, the RL agent explores conformations where the radius of gyration falls within the target range $targetRG_{min}$ = 0 [nm], $targetRG_{max}$ = 20 [nm]. Consequently, conformations within the target range were sampled more frequently, leading to a higher concentration of conformations $\rho_\pi(\tau)$ around the target radius of gyration, resulting in desired trajectories. The probability of the trajectory $\tau$ of the polymer chain can be described based on the expression:
\[
\rho_{\pi}(\tau) = p(s_0) \prod_{t=0}^{T-1} \pi(f_{t\text{ learned}}|s_t) p(s_{t+1}|s_t,f_{t\text{ learned}})  \tag{2}
\]
In expression (2), $p(s_0)$ represents the initial state $s_0$ probability, which captures the likelihood of the polymer chain starting in a particular conformation. This probability depends on the initial conditions and can influence the exploration of different regions of the conformational space. The term $\pi(f_{t\text{ learned}}|s_t)$ represents the policy $\pi$ that selects the action $f_{t\text{ learned}}$ given the current state $s_t$. The policy $\pi$ is responsible for shaping the trajectory of the polymer chain by guiding the selection of actions that maximize expected rewards.

\textbf{Transition $T_t$: } The transition probability $T_t$ defined by $p(s_{t+1}|s_t,f_{t\text{ learned}})$ which captures the likelihood of transitioning from the current state $s_t$ to the next state $s{t+1}$ given the reward $r_t$ and the applied force $f_{t\text{ learned}}$. This probability reflects how the dynamics of the polymer chain, influenced by the applied forces and the rewards obtained, lead to transitions between different conformations. The probability $\rho_{\pi}(\tau)$ represents the observability of the trajectory $\tau$ of the polymer chain under the policy $\pi$. It captures the likelihood of traversing a sequence of states as determined by the policy $\pi$ optimized by the P5 model.

\section{Results}

Our study focused on Cellulose Acetate, a renewable and biodegradable biopolymer. The cellulose acetate molecule typically has between 65 to 350 monomers and is represented by a coarse-grained model following the Martini Force Field with 75 monomers of 458 [g/mol]. The P5 model was trained for approximately 35 million episodes with 20,000 simulation steps each. The trained policy was then implemented in real-time simulation, enabling continuous adjustment to ensure the polymer chain remained within the target RG range of 0 to 20 nanometers throughout the runtime. During the training phase, the cumulative reward versus the environment optimized the policy quickly, as shown in Figure 2. Intermittent dips in the cumulative reward were attributed to the potential energy excess induced by initially compressed conformations and high potential energy spikes. Despite these dips, the NN converged with sustained high values after six hours of training on a conventional laptop. The P5 model demonstrates the effectiveness of machine learning in efficiently controlling the conformational behavior of the Cellulose Acetate polymer. 

\begin{figure}[h!]
    \centering
    \includegraphics[width=0.47\textwidth]{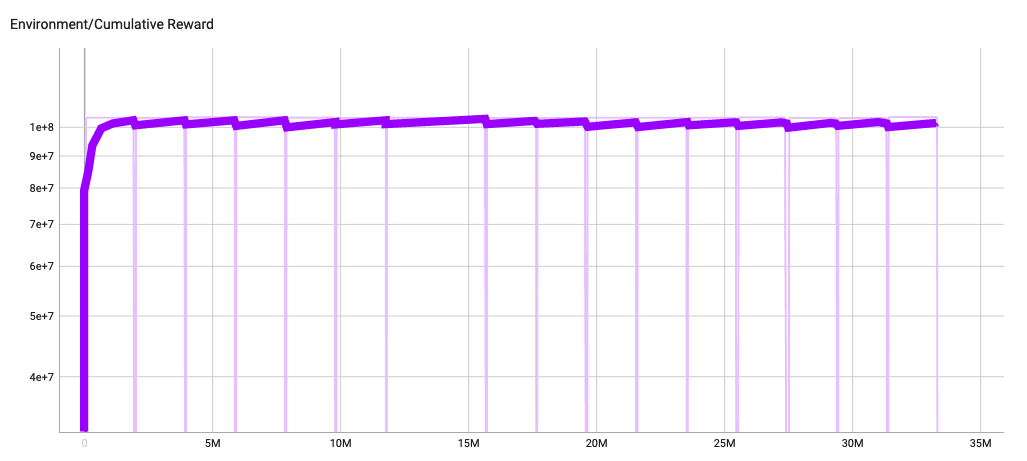}

    \caption{Learning curve used to monitor training progress.}
\end{figure}

Our results demonstrate the effectiveness of the P5 model in efficiently controlling the conformational behavior of a simulated molecule of Cellulose Acetate. With optimized policies, P5 successfully guides the polymer chain to stay within the target range of the RG, resulting in a $37.1\%$ increase in sampling targeted states, as depicted in Figure 3. P5 effectively corrected conformation spikes due to high initial potential energy. Notably, P5 achieves these target conformations in half the wall clock time than MD. 

\begin{figure}[ht]
    \centering
    \includegraphics[width=0.47\textwidth]{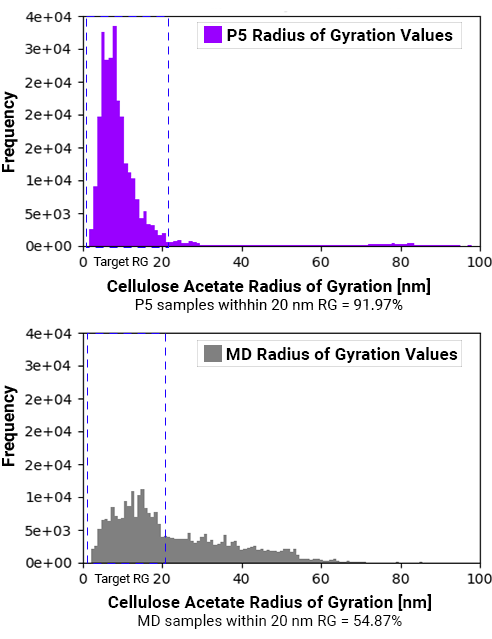}

    \caption{Comparing Conformations Control: P5 Model vs. Molecular Dynamics Simulations in Cellulose Acetate.}
\end{figure}

The histogram trajectories shown in Figure 4, generated by P5, demonstrate the robustness and effectiveness of the model in correcting and guiding the polymer chain's conformation toward the targeted RG values. By manually changing the initial state and starting from different RG values, the policy network within P5 dynamically adjusts the forces acting on the beads to ensure the trajectory aligns with the desired RG range. This showcases the model's ability to adapt to unseen states and highlights its computational viability and transferability to other polymers and potentially de Novo molecules.

\begin{figure}[h!]
    \centering
    \includegraphics[width=0.45\textwidth]{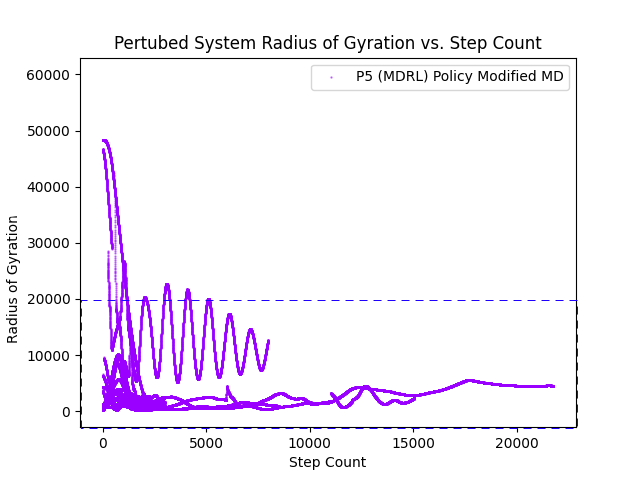}
    \caption{ Trajectory Corrections of Polymer Chain Conformations by P5 Model from Unseen States (manually perturbed initial states explained in Appendix 4 and 5).}
\end{figure}

\section{Contributions and Future work}

In this work, we introduced the P5 model, a powerful tool that leverages machine learning control policies to optimize molecular dynamics simulations and expand their generative capabilities. The P5 simulator's unique integration of enhanced optimization, computational efficiency, and intuitive usability offers a novel approach for property evaluation, morphology analysis, and mechanical property prediction in complex molecular systems. Regarding efficiency and cost-effectiveness, P5 significantly outperforms traditional MD simulations by close to $40\%$, marking a pivotal step towards the computational design of molecular systems using generative machine learning techniques. We plan to extend the P5 model to a broader range of polymer structures and full-atom simulations, providing a more comprehensive understanding of the model's efficacy and scalability. Additionally, we aim to explore innovative control algorithms and integrate other physics machine learning models, analytical approaches, and time-coarse optimization to enhance the P5 model's performance further. These research paths are promising approaches that combine physics-based machine learning and molecular dynamics simulations.

\clearpage
\bibliography{referencesBibText}

\begin{thebibliography}{22}
\providecommand{\natexlab}[1]{#1}
\providecommand{\url}[1]{\texttt{#1}}
\expandafter\ifx\csname urlstyle\endcsname\relax
  \providecommand{\doi}[1]{doi: #1}\else
  \providecommand{\doi}{doi: \begingroup \urlstyle{rm}\Url}\fi

\bibitem[Antoniuk et~al.(2022)Antoniuk, Li, Kailkhura, and
  Hiszpanski]{antoniuk_representing_2022}
Antoniuk, E.~R., Li, P., Kailkhura, B., and Hiszpanski, A.~M.
\newblock Representing {Polymers} as {Periodic} {Graphs} with {Learned}
  {Descriptors} for {Accurate} {Polymer} {Property} {Predictions}, May 2022.
\newblock URL \url{http://arxiv.org/abs/2205.13757}.
\newblock arXiv:2205.13757 [cond-mat].

\bibitem[Bereau \& Kremer(2015)Bereau and Kremer]{bereau_automated_2015}
Bereau, T. and Kremer, K.
\newblock Automated {Parametrization} of the {Coarse}-{Grained} {Martini}
  {Force} {Field} for {Small} {Organic} {Molecules}.
\newblock \emph{Journal of Chemical Theory and Computation}, 11\penalty0
  (6):\penalty0 2783--2791, June 2015.
\newblock ISSN 1549-9618.
\newblock \doi{10.1021/acs.jctc.5b00056}.
\newblock URL \url{https://doi.org/10.1021/acs.jctc.5b00056}.
\newblock Publisher: American Chemical Society.

\bibitem[Brockman et~al.(2016)Brockman, Cheung, Pettersson, Schneider,
  Schulman, Tang, and Zaremba]{brockman_openai_2016}
Brockman, G., Cheung, V., Pettersson, L., Schneider, J., Schulman, J., Tang,
  J., and Zaremba, W.
\newblock {OpenAI} {Gym}, June 2016.
\newblock URL \url{http://arxiv.org/abs/1606.01540}.
\newblock arXiv:1606.01540 [cs].

\bibitem[Doerr et~al.(2021)Doerr, Majewski, Pérez, Krämer, Clementi, Noe,
  Giorgino, and De~Fabritiis]{doerr_torchmd_2021}
Doerr, S., Majewski, M., Pérez, A., Krämer, A., Clementi, C., Noe, F.,
  Giorgino, T., and De~Fabritiis, G.
\newblock {TorchMD}: {A} {Deep} {Learning} {Framework} for {Molecular}
  {Simulations}.
\newblock \emph{Journal of Chemical Theory and Computation}, 17\penalty0
  (4):\penalty0 2355--2363, April 2021.
\newblock ISSN 1549-9618.
\newblock \doi{10.1021/acs.jctc.0c01343}.
\newblock URL \url{https://doi.org/10.1021/acs.jctc.0c01343}.
\newblock Publisher: American Chemical Society.

\bibitem[Ganea et~al.(2021)Ganea, Pattanaik, Coley, Barzilay, Jensen, Green,
  and Jaakkola]{ganea_geomol_2021}
Ganea, O.-E., Pattanaik, L., Coley, C.~W., Barzilay, R., Jensen, K.~F., Green,
  W.~H., and Jaakkola, T.~S.
\newblock {GeoMol}: {Torsional} {Geometric} {Generation} of {Molecular} {3D}
  {Conformer} {Ensembles}, June 2021.
\newblock URL \url{http://arxiv.org/abs/2106.07802}.
\newblock Number: arXiv:2106.07802 arXiv:2106.07802 [physics].

\bibitem[Holdijk et~al.(2022)Holdijk, Du, Hooft, Jaini, Ensing, and
  Welling]{holdijk_path_2022}
Holdijk, L., Du, Y., Hooft, F., Jaini, P., Ensing, B., and Welling, M.
\newblock Path {Integral} {Stochastic} {Optimal} {Control} for {Sampling}
  {Transition} {Paths}, June 2022.
\newblock URL \url{http://arxiv.org/abs/2207.02149}.
\newblock arXiv:2207.02149 [physics, q-bio].

\bibitem[Juliani et~al.(2020)Juliani, Berges, Teng, Cohen, Harper, Elion, Goy,
  Gao, Henry, Mattar, and Lange]{juliani_unity_2020}
Juliani, A., Berges, V.-P., Teng, E., Cohen, A., Harper, J., Elion, C., Goy,
  C., Gao, Y., Henry, H., Mattar, M., and Lange, D.
\newblock Unity: {A} {General} {Platform} for {Intelligent} {Agents}, May 2020.
\newblock URL \url{http://arxiv.org/abs/1809.02627}.
\newblock arXiv:1809.02627 [cs, stat].

\bibitem[Karuth et~al.(2021)Karuth, Alesadi, Xia, and
  Rasulev]{karuth_predicting_2021}
Karuth, A., Alesadi, A., Xia, W., and Rasulev, B.
\newblock Predicting glass transition of amorphous polymers by application of
  cheminformatics and molecular dynamics simulations.
\newblock \emph{Polymer}, 218:\penalty0 123495, March 2021.
\newblock ISSN 0032-3861.
\newblock \doi{10.1016/j.polymer.2021.123495}.
\newblock URL
  \url{https://www.sciencedirect.com/science/article/pii/S003238612100118X}.

\bibitem[Klein et~al.(2023)Klein, Foong, Fjelde, Mlodozeniec, Brockschmidt,
  Nowozin, Noé, and Tomioka]{klein_timewarp_2023}
Klein, L., Foong, A. Y.~K., Fjelde, T.~E., Mlodozeniec, B., Brockschmidt, M.,
  Nowozin, S., Noé, F., and Tomioka, R.
\newblock Timewarp: {Transferable} {Acceleration} of {Molecular} {Dynamics} by
  {Learning} {Time}-{Coarsened} {Dynamics}, February 2023.
\newblock URL \url{http://arxiv.org/abs/2302.01170}.
\newblock arXiv:2302.01170 [cond-mat, physics:physics, stat].

\bibitem[Kleinschmidt et~al.(2022)Kleinschmidt, Chen, Pascal, and
  Lipomi]{kleinschmidt_computational_2022}
Kleinschmidt, A.~T., Chen, A.~X., Pascal, T.~A., and Lipomi, D.~J.
\newblock Computational {Modeling} of {Molecular} {Mechanics} for the
  {Experimentally} {Inclined}.
\newblock \emph{Chemistry of Materials}, 34\penalty0 (17):\penalty0 7620--7634,
  September 2022.
\newblock ISSN 0897-4756.
\newblock \doi{10.1021/acs.chemmater.2c00292}.
\newblock URL \url{https://doi.org/10.1021/acs.chemmater.2c00292}.
\newblock Publisher: American Chemical Society.

\bibitem[Landrum(2010)]{landrum_rdkit_2010}
Landrum, G.
\newblock {RDKit}, 2010.
\newblock URL \url{https://www.rdkit.org/}.

\bibitem[Liu et~al.(2023)Liu, Luo, Xu, Jaakkola, and Tegmark]{liu_genphys_2023}
Liu, Z., Luo, D., Xu, Y., Jaakkola, T., and Tegmark, M.
\newblock {GenPhys}: {From} {Physical} {Processes} to {Generative} {Models},
  April 2023.
\newblock URL \url{http://arxiv.org/abs/2304.02637}.
\newblock arXiv:2304.02637 [physics, physics:quant-ph].

\bibitem[Mnih et~al.(2016)Mnih, Badia, Mirza, Graves, Lillicrap, Harley,
  Silver, and Kavukcuoglu]{mnih_asynchronous_2016}
Mnih, V., Badia, A.~P., Mirza, M., Graves, A., Lillicrap, T.~P., Harley, T.,
  Silver, D., and Kavukcuoglu, K.
\newblock Asynchronous {Methods} for {Deep} {Reinforcement} {Learning}, June
  2016.
\newblock URL \url{http://arxiv.org/abs/1602.01783}.
\newblock arXiv:1602.01783 [cs].

\bibitem[Muratore et~al.(2022)Muratore, Ramos, Turk, Yu, Gienger, and
  Peters]{muratore_robot_2022}
Muratore, F., Ramos, F., Turk, G., Yu, W., Gienger, M., and Peters, J.
\newblock Robot {Learning} {From} {Randomized} {Simulations}: {A} {Review}.
\newblock \emph{Frontiers in Robotics and AI}, 9, 2022.
\newblock ISSN 2296-9144.
\newblock URL
  \url{https://www.frontiersin.org/articles/10.3389/frobt.2022.799893}.

\bibitem[Park et~al.(2022)Park, Shim, Lee, Rammohan, Goyal, Shim, Jeong, and
  Kim]{park_prediction_2022}
Park, J., Shim, Y., Lee, F., Rammohan, A., Goyal, S., Shim, M., Jeong, C., and
  Kim, D.~S.
\newblock Prediction and {Interpretation} of {Polymer} {Properties} {Using} the
  {Graph} {Convolutional} {Network}.
\newblock \emph{ACS Polymers Au}, January 2022.
\newblock \doi{10.1021/acspolymersau.1c00050}.
\newblock URL \url{https://doi.org/10.1021/acspolymersau.1c00050}.
\newblock Publisher: American Chemical Society.

\bibitem[Rose et~al.(2021)Rose, Mair, and Garrahan]{rose_reinforcement_2021}
Rose, D.~C., Mair, J.~F., and Garrahan, J.~P.
\newblock A reinforcement learning approach to rare trajectory sampling.
\newblock \emph{New Journal of Physics}, 23\penalty0 (1):\penalty0 013013,
  January 2021.
\newblock ISSN 1367-2630.
\newblock \doi{10.1088/1367-2630/abd7bd}.
\newblock URL \url{https://dx.doi.org/10.1088/1367-2630/abd7bd}.
\newblock Publisher: IOP Publishing.

\bibitem[Schulman et~al.(2017)Schulman, Wolski, Dhariwal, Radford, and
  Klimov]{schulman_proximal_2017}
Schulman, J., Wolski, F., Dhariwal, P., Radford, A., and Klimov, O.
\newblock Proximal {Policy} {Optimization} {Algorithms}, August 2017.
\newblock URL \url{http://arxiv.org/abs/1707.06347}.
\newblock arXiv:1707.06347 [cs].

\bibitem[Tao et~al.(2021)Tao, Varshney, and Li]{tao_benchmarking_2021}
Tao, L., Varshney, V., and Li, Y.
\newblock Benchmarking {Machine} {Learning} {Models} for {Polymer}
  {Informatics}: {An} {Example} of {Glass} {Transition} {Temperature}.
\newblock \emph{Journal of Chemical Information and Modeling}, 61\penalty0
  (11):\penalty0 5395--5413, November 2021.
\newblock ISSN 1549-9596.
\newblock \doi{10.1021/acs.jcim.1c01031}.
\newblock URL \url{https://doi.org/10.1021/acs.jcim.1c01031}.
\newblock Publisher: American Chemical Society.

\bibitem[Todorov et~al.(2012)Todorov, Erez, and Tassa]{todorov_mujoco_2012}
Todorov, E., Erez, T., and Tassa, Y.
\newblock {MuJoCo}: {A} physics engine for model-based control.
\newblock In \emph{2012 {IEEE}/{RSJ} {International} {Conference} on
  {Intelligent} {Robots} and {Systems}}, pp.\  5026--5033, October 2012.
\newblock \doi{10.1109/IROS.2012.6386109}.
\newblock ISSN: 2153-0866.

\bibitem[Unke et~al.(2021)Unke, Chmiela, Sauceda, Gastegger, Poltavsky,
  Schütt, Tkatchenko, and Müller]{unke_machine_2021}
Unke, O.~T., Chmiela, S., Sauceda, H.~E., Gastegger, M., Poltavsky, I.,
  Schütt, K.~T., Tkatchenko, A., and Müller, K.-R.
\newblock Machine {Learning} {Force} {Fields}.
\newblock \emph{Chemical Reviews}, 121\penalty0 (16):\penalty0 10142--10186,
  August 2021.
\newblock ISSN 0009-2665.
\newblock \doi{10.1021/acs.chemrev.0c01111}.
\newblock URL \url{https://doi.org/10.1021/acs.chemrev.0c01111}.
\newblock Publisher: American Chemical Society.

\bibitem[You et~al.(2018)You, Liu, Ying, Pande, and Leskovec]{you_graph_2018}
You, J., Liu, B., Ying, Z., Pande, V., and Leskovec, J.
\newblock Graph {Convolutional} {Policy} {Network} for {Goal}-{Directed}
  {Molecular} {Graph} {Generation}.
\newblock In \emph{Advances in {Neural} {Information} {Processing} {Systems}},
  volume~31. Curran Associates, Inc., 2018.
\newblock URL
  \url{https://proceedings.neurips.cc/paper/2018/hash/d60678e8f2ba9c540798ebbde31177e8-Abstract.html}.

\bibitem[Zuo et~al.(2021)Zuo, Qin, Chen, Ye, Li, Luo, and
  Ong]{zuo_accelerating_2021}
Zuo, Y., Qin, M., Chen, C., Ye, W., Li, X., Luo, J., and Ong, S.~P.
\newblock Accelerating {Materials} {Discovery} with {Bayesian} {Optimization}
  and {Graph} {Deep} {Learning}, April 2021.
\newblock URL \url{http://arxiv.org/abs/2104.10242}.
\newblock arXiv:2104.10242 [cond-mat].

\end{thebibliography}
\bibliographystyle{icml2023}

\newpage
\appendix
\onecolumn
\section{Appendix}

\subsection{Simulation Architecture}
The development of the P5 model was a complex endeavor that required integrating a molecular dynamics (MD) module within Unity3D, a robust physics engine, and the inclusion of reinforcement learning (RL). This project's complexity was heightened due to the need for accurate representation and simulation of atomic and molecular interactions, including Brownian random kicks that eventually evolved into a learned force within the system. Unity3D facilitated the intricate modeling and real-time visual inspection of simulations, which is critical for verifying model accuracy. Meanwhile, the integration of RL demanded the creation of a high-performing reward function and a balance between exploration and exploitation trade-offs. Amid these challenges, we opted for Proximal Policy Optimization (PPO) for its robustness, computational efficiency, and relative simplicity, aiding more straightforward integration with the MD module. Furthermore, despite the lack of extensive precedents, our pioneering effort in developing control policies for molecular dynamics contributed to establishing a robust system, setting the stage for future refinement and optimization.

Quantifying the effects and efficiency of the reinforcement learning-driven Brownian "kicks" in P5 molecular dynamics (MD) simulations is crucial. Unity 3D visualization plays a significant role in this process by providing an additional layer of validation, contributing to the robustness of the simulations. The raw data generated from MD simulations relevant for P5 include particle coordinates, velocity, and acceleration data, energy data (kinetic and potential energies), bond lengths, angles, dihedrals, and diffusion coefficients. This information helps track particles' motion and energy changes influenced by the Brownian "kicks" and identify structural changes or bond breakages that could result from these rapid perturbations.

Despite the comprehensiveness of this raw data, it might overlook certain rapid phenomena, such as bond breakages induced by Brownian "kicks." These events occur so swiftly that their detection might be challenging when solely relying on raw data analysis. This error detection is where Unity 3D visualization becomes invaluable, providing an interactive, visually rich platform for identifying and rectifying inconsistencies or errors. By making potential issues visible that the raw data may fail to detect, Unity 3D visualization ensures the simulations' accuracy and validity, proving indispensable in the analytical toolkit.

\subsection{Timescale Factor Calculation}
To convert the dimensionless timescale to physical units, specifically milliseconds, we use the fundamental relationship $\tau = \frac{r}{v}$, where $\tau$ is the characteristic time, $r$ is the characteristic length, and $v$ is the characteristic velocity.

    Characteristic Length ($r$): In the P5 model, the units of length are in Angstroms (Å), which equals $10^{-10}$ m.

    Characteristic Mass ($m$): The model uses mass units based on the Van der Waals mass. For a polymer composed of monomers with seven beads, each with a total molecular weight of 458 g/mol, the characteristic mass of each bead ($m_{\text{bead}}$) is calculated by dividing this molecular weight by 7 (to account for the seven beads), and by Avogadro's number $N_A = 6.022 \times 10^{23}$ to convert to atomic mass units.

    Characteristic Velocity ($v$): The characteristic velocity comes from the kinetic theory of gases and the equipartition theorem, which states that the mean kinetic energy of a molecule due to its motion through space is given by $\frac{1}{2} m v^2 = \frac{3}{2} kT$, where $m$ is mass, $v$ is velocity, $k$ is Boltzmann's constant, and $T$ is temperature. Here, we can calculate the characteristic velocity $v_{\text{char}}$ when $T$ is set to a typical room temperature (say, 298 K) and $m$ is set to $m_{\text{bead}}$.

    Characteristic Time ($\tau$): The characteristic time can then be found by substituting $r$ and $v_{\text{char}}$ into the equation $\tau = \frac{r}{v_{\text{char}}}$.

    The timestep factor to convert from dimensionless to femtoseconds is 209.7915273799608.

\subsection{Martnini Model and Hyperparameters}

This model results in seven types of beads, each categorized as "Na" (non-polar, hydrogen-acceptor), "P3" (polar, polarity level 3), or "SP1" (small, in ring, polar, polarity level 1). These beads carry assigned masses according to the Martini Model. Van der Waals' diameters are 5.2 Angstroms for regular beads and 4.7 Angstroms for smaller beads in a ring (designated with an 'S' prefix). Additionally, we incorporate an implicit solvent model in the form of Brownian dynamics to provide a more representative simulation environment, accounting for the effects of the solvent on the behavior of our molecular system.

\begin{figure}[h!]
\vskip 0.2in
\begin{center}
\centerline{\includegraphics[width=0.5\columnwidth]{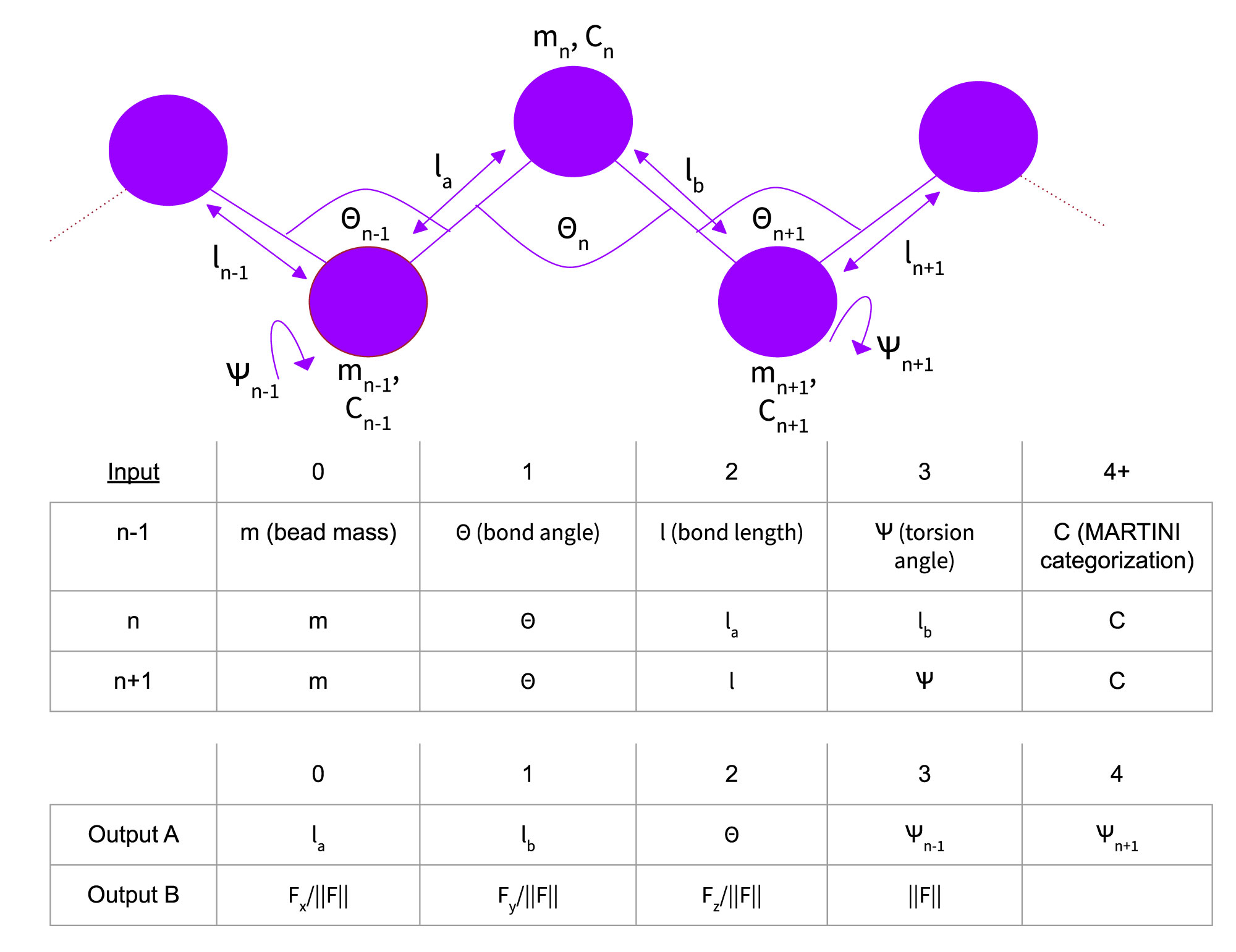}}
\caption{Parameters of the polymer chain agent.}
\label{icml-historical}
\end{center}
\vskip -0.2in
\end{figure}

\subsection{Generative P5}
We conducted a series of experiments to demonstrate the generative capabilities of the P5 model. We manually perturbed the initial state of 10 episodes, each consisting of 20,000 steps, while keeping the P5-MD system in standby mode. By changing the positions of the beads in the initial state, we introduced variations in the starting configurations. Remarkably, the P5 model effectively corrected the trajectory in each episode, guiding the polymer chain toward the desired gyration (RG) range radius. This showcases the model's ability to generate new conformations within a target space by leveraging its learned control policies. These generated conformations serve as valuable data points that can be further analyzed to gain insights into specific functional groups' structural properties, energy landscape, and behavior. Moreover, these conformations can be utilized in various downstream applications, such as predicting molecular properties, exploring structure-activity relationships, or conducting virtual screening for drug discovery. The P5 model's generative capabilities open up exciting opportunities for researchers to expand their understanding of molecular behavior, explore previously uncharted regions of chemical space, and drive innovation in diverse scientific domains.

Further, P5 autonomously generates data through modified molecular dynamics simulations, obviating the reliance on substantial experimental chemical datasets typically required in similar approaches. P5 finds conformations that may not be accessible to pure molecular dynamics.

A.5. compares gyration radius (RG) trajectories from P5 and traditional molecular dynamics (MD) simulations that provide critical insights into the conformational latent space. RG trajectories from pure MD showcase the molecule's inherent behavior. In contrast, P5 trajectories, driven by reinforcement learning-guided Brownian "kicks," demonstrate an enhanced ability to explore broader or alternative regions within this latent space. This comparison reveals the influence and effectiveness of the learned "kicks" in navigating the latent space, serving as a pivotal tool in assessing P5's generative capabilities and directing future model improvements.

\clearpage
\subsection{Trajectories Comparison}

\begin{figure}[h!]
    \centering
    \includegraphics[width=0.7\textwidth]{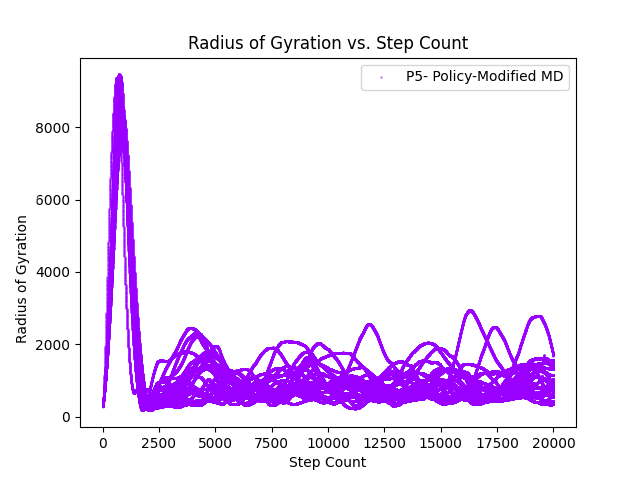}
    \includegraphics[width=0.7\textwidth]{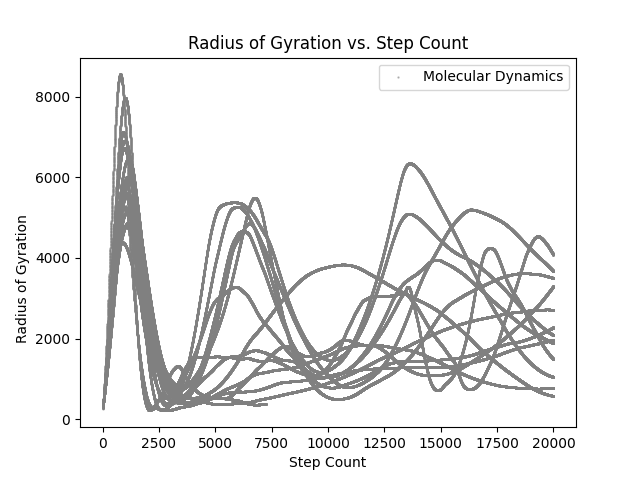}
    \caption{P5 \& MD trajectories for the Radius of Gyration Values.}
\end{figure}


\end{document}